\documentclass[letterpaper]{article} 
\usepackage{aaai24}  
\usepackage{times}  
\usepackage{helvet}  
\usepackage{courier}  
\usepackage[hyphens]{url}  
\usepackage{graphicx} 
\usepackage{natbib}  
\usepackage{caption} 
\usepackage{natbib}
\setcitestyle{numbers,square}
\usepackage[utf8]{inputenc} 
\usepackage[T1]{fontenc}    
\usepackage{url}            
\usepackage{booktabs}       
\usepackage{amsfonts}       
\usepackage{nicefrac}       
\usepackage{microtype}      
\usepackage{xcolor}         

\usepackage{graphicx} 
\usepackage{algpseudocode}
\usepackage{algorithm}  
\usepackage{amsmath}

\usepackage{newfloat}
\usepackage{listings}
\DeclareCaptionStyle{ruled}{labelfont=normalfont,labelsep=colon,strut=off} 
\floatstyle{ruled}
\newfloat{listing}{tb}{lst}{}
\floatname{listing}{Listing}
\title{LCV2I: Communication-Efficient and High-Performance Collaborative \\ Perception Framework with Low-Resolution LiDAR}
\author{
    Xinxin Feng,
    Haoran Sun,
    HaiFeng Zheng
    \thanks{Corresponding author}
   }
   \affiliations{
     Fujian Key Lab for Intelligent Processing and Wireless Transmission of Media Information,\\
     College of Physics and Information Engineering, Fuzhou University, Fuzhou 35108, China\\
    E-mails:\{fxx1116,221127183,zhenghf\}@fzu.edu.cn
}
\usepackage{bibentry}

\begin{document}

\maketitle

\begin{abstract}

Vehicle-to-Infrastructure (V2I) collaborative perception leverages data collected by infrastructure's sensors to enhance vehicle perceptual capabilities.  
LiDAR, as a commonly used sensor in cooperative perception, is widely equipped in intelligent vehicles and infrastructure. However, its superior performance comes with a correspondingly high cost. To achieve low-cost V2I, reducing the cost of LiDAR is crucial. 
Therefore, we study adopting low-resolution LiDAR on the vehicle to minimize cost as much as possible.
However, simply reducing the resolution of vehicle's LiDAR results in sparse point clouds, making distant small objects even more blurred. Additionally, traditional communication methods have relatively low bandwidth utilization efficiency. These factors pose challenges for us.
To balance cost and perceptual accuracy, we propose a new collaborative perception framework, namely LCV2I.  
LCV2I uses data collected from cameras and low-resolution LiDAR as input. It also employs feature offset correction modules and regional feature enhancement algorithms to improve feature representation. Finally, we use regional difference map and regional score map to assess the value of collaboration content, thereby improving communication bandwidth efficiency.
In summary, our approach achieves high perceptual performance while substantially reducing the demand for high-resolution sensors on the vehicle. To evaluate this algorithm, we conduct 3D object detection in the real-world scenario of DAIR-V2X, demonstrating that the performance of LCV2I consistently surpasses currently existing algorithms.
  
\end{abstract}

\section{Introduction}

The emergence of collaborative sensing technology has resolved the limitations of individual intelligent vehicles in effectively detecting distant small targets and occluded objects.
Currently, commonly used collaborative perception technologies include Vehicle-to-Infrastructure (V2I) and Vehicle-to-Vehicle (V2V). As the names suggest, V2I involves deploying sensors on the infrastructure to assist vehicle driving, while V2V utilizes sensors on surrounding vehicles to assist the ego vehicle. Each method has its pros and cons, but V2I, with sensors often deployed at elevated positions on the infrastructure, allows the sensors to assist intelligent vehicles from a higher perspective, providing a broader view to better sense the surrounding environment. This helps overcome the limitations of the single perspective of intelligent vehicles. 

Based on existing V2X simulation datasets such as DAIR-V2X \cite{9879243}, OpenV2V \cite{opv2v}, V2X-Sim \cite{9294660}, and V2XSet \cite{10.1007/978-3-031-19842-7_7}, previous researchers have made significant contributions. Where2comm \cite{Where2comm:22} introduces a collaborative approach based on a spatial confidence map to reduce transmission bandwidth. V2VNet \cite{10.1007/978-3-030-58536-5_36} explores end-to-end learning with source code encoding, and DiscoNet \cite{NEURIPS2021_f702defb} uses 1D convolution to compress messages.  
However, it's important to note that all these approaches were developed under the assumption of high-resolution LiDAR sensors on both vehicles and infrastructure. While high-resolution LiDAR provides higher detection accuracy, it also increases unnecessary costs.

To address the issue mentioned above, we consider reducing the resolution of the vehicle's LiDAR by lowering the number of scan lines of LiDAR to half or even one-fourth of the original, while keeping the infrastructure LiDAR's resolution unchanged to provide an auxiliary field of view. However, we will face the following four issues: Firstly, LiDAR has poor perception capability for distant small targets, and this weakness is more pronounced with low-resolution LiDAR, thus requiring additional modality data for supplementation.
Secondly, due to the different fields of view of various sensors for the same area, even though data from different sensors have been projected onto a unified coordinate system, there will still be areas only sensed by a single sensor. 
Therefore, during the fusion process, some areas will be well fused, while others will be insufficiently fused.
Thirdly, as the point cloud becomes sparser, the regional features of the target will become blurred, leading to increased false detections and missed detections.
Finally, since most current collaborative strategies send all auxiliary information to the vehicle, the utilization of communication bandwidth is extremely low.

To address the aforementioned issues, we propose a Low-Cost and high-performance V2I collaborative perception framework (LCV2I), which adopts multi-modal data as input. This framework primarily consists of three key modules:
(1) Voxel-Wise Fusion Module (VWF), which uses pixel features from camera data to complement voxel features on a voxel-by-voxel basis.
(2) Feature Offset Correction Module (FOCM), which adjusts the feature weights offset after cross-modal fusion.
(3) Regional Feature Enhancement Algorithm (RFEA), which enhances the region features of the target to be perceived. Simultaneously, RFEA generates a feature difference matrix to guide the generation of regional difference map and optimize the region scoring map to reduce transmission bandwidth.
The main contributions of our work are as follows:
\begin{itemize}
    \item Considering the high cost of vehicles' sensors, we propose LCV2I, a low-cost, high-bandwidth-utilization, and high-performance collaborative perception framework. LCV2I allows vehicles to be equipped with low-resolution sensors to achieve high-performance 3D object detection, while only transmitting critical features, thereby ensuring high bandwidth utilization.
    \item We propose VWF and FOCM to utilize image features to supplement voxel features, addressing the issue of inaccurate voxel feature generation caused by the reduced resolution of LiDAR.
    \item We propose RFEA to enhance the regional features of targets while generating a regional difference map to guide the filtering of invalid features during communication. This further improves communication bandwidth utilization, achieving a balance between bandwidth and performance.
    \item To evaluate LCV2I, we conducted a collaborative 3D object detection task on the real-world dataset DAIR-V2X \cite{9879243}. The results indicate that our approach maintains high perceptual accuracy and low transmission bandwidth, even with a significant reduction in sensor resolution.
\end{itemize}

\section{Related Work}

\subsection{Collaborative Perception}
 Typically, collaborative perception involves fusing information perceived by infrastructure's sensors with vehicle's sensor perception information to enhance the global perception capability of the vehicle's. Based on different collaboration stages, previous work can be broadly categorized into early, intermediate, and late collaboration.
\subsubsection{Early Collaboration and Late Collaboration}
Cooper \cite{8885377} primarily shares multi-precision LiDAR points, projecting its sparse representation into a compact space, followed by a sparse point cloud object detection network to adapt to low-density point clouds. However, early fusion comes with significant computational overhead. In contrast, late fusion involves directly communicating and fusing perception results from different intelligent agents. TruPercept \cite{9304695} introduces a trust mechanism for secure message selection, adjusting based on the trust model of the intelligent agent providing perception results. It enhances perception performance beyond the line of sight and at a distance from the vehicle by fusing communication messages, relying on local-verified reports of object detection accuracy. However, late fusion is subjective, and once the perception of a collaborator is compromised, the impact is also shared by the collaborator.

\subsubsection{Intermediate Collaboration}
To strike a balance between perception accuracy and reasoning latency, intermediate fusion methods have been widely explored. V2VNet proposes a graph-based approach, capturing and updating the geographical information of each vehicle iteratively through Convolutional Gated Recurrent Units (ConvGRU). To emphasize the importance of agents, DiscoNet discards highly similar pixels between vehicles through an edge weight matrix and constructs an overall geometric topology through knowledge distillation. To simulate the impact of real-world transmission delays, Who2com \cite{9197364} proposes a three-step handshake communication protocol, including request, match, and connect, to determine which collaborator to interact with. Additionally, When2com \cite{Liu_2020_CVPR} considers a learnable self-attention mechanism to infer whether the self-agent engages in additional communication for more information. Where2comm develops a novel sparse confidence map to mask unimportant elements used for feature compression. Investigating fine-grained and dense predictions for in-vehicle cameras, CoBEVT \cite{pmlr-v205-xu23a} studies a pure camera map prediction framework below the Bird's Eye View (BEV) plane. This framework utilizes a novel Fusion Axis (FAX) attention to reconstruct dynamic scenes on the ground plane.

Despite the excellent performance of the above algorithms, they are all built on the foundation of high-resolution sensors, without considering the cost issues associated with practical deployment. In this work, we propose a V2I framework that leverages low-resolution sensors to achieve high-precision perception. We further optimize the intermediate features to achieve more efficient bandwidth utilization.

\begin{figure*}[h]
	\centering
	\includegraphics[height=85mm]{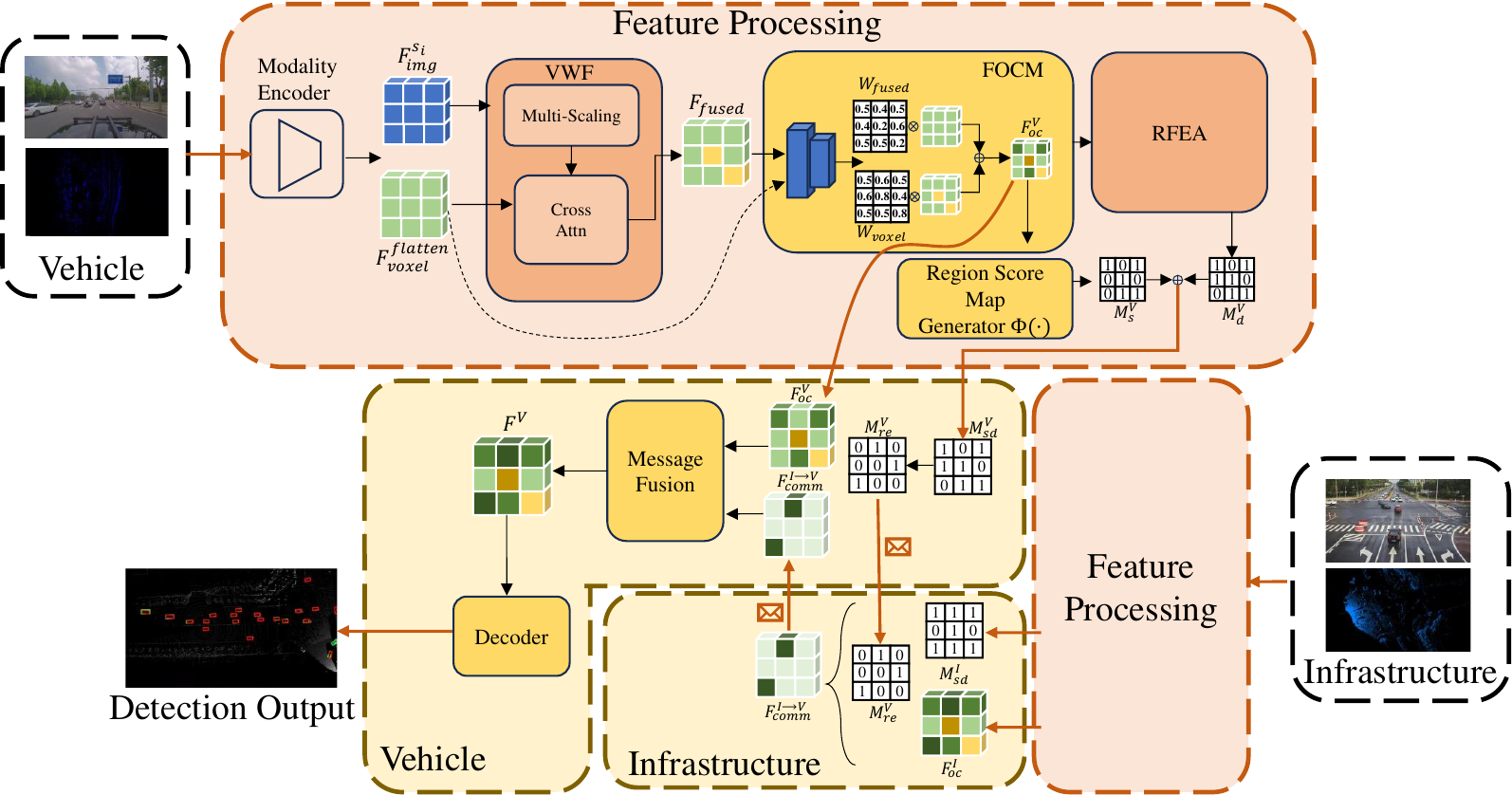}
	\caption{System architecture diagram of LCV2I}
	\label{fig: sys}
\end{figure*}

\section{Method}

LCV2I is applied in vehicle-road collaborative perception scenarios, enabling better perception results with the assistance of infrastructure even when the vehicle's LiDAR resolution is significantly reduced. The system architecture of LCV2I is shown in Fig. \ref{fig: sys}. Both the vehicle and infrastructure use the same network structure, consisting of the following components: Modality Encoder, VWF, FOCM, RFEA, and Collaborative Perception Module. The VWF module uses image features to supplement the missing feature information from the low-resolution LiDAR, the FOCM module corrects feature offsets generated during fusion, the RFEA module enhances the regional characteristics of features and generates a regional difference map, and the Collaboration Module uses the regional difference map to achieve higher communication bandwidth utilization.

\subsection{Modality Encoder}
The modality encoder extracts features from sensor data. In LCV2I, it uses two modalities of data, namely LiDAR point clouds and RGB images, as inputs. For the LiDAR point cloud data collected from the LiDAR sensor, it is voxelized and then encoded using the PointPillars \cite{8954311}. For the RGB images captured by the camera, they are processed using ResNet50 \cite{he2016deep} to extract features. Additionally, the features are represented from a Bird's Eye View (BEV) perspective, and the perception information from all intelligent agents is projected onto a unified global coordinate system to facilitate better collaboration and avoid complex coordinate transformations.

For the input RGB image $X_c$ and point cloud $X_l$, we can extract their features $F_{img}^{ori}=F_{enc}(X_c)\in\mathbb{R}^{b\times c\times {h_c}\times {w_c}}$ and $F_{voxel}^{ori}=F_{enl}(voxelize(X_l))\in\mathbb{R}^{b\times c\times z\times h\times w}$, where ${F_{enc}}$ and ${F_{enl}}$ are the image feature encoder and voxel feature encoder, respectively, and $b$, $c$, $z$, $h$, $w$, $h_c$ and $w_c$ represent the batch size, number of channels, height, length, and width of voxel features, as well as height and width of image features.
\subsection{Voxel-Wise Fusion module}
\begin{figure}
	\centering
	\includegraphics[height=55mm]{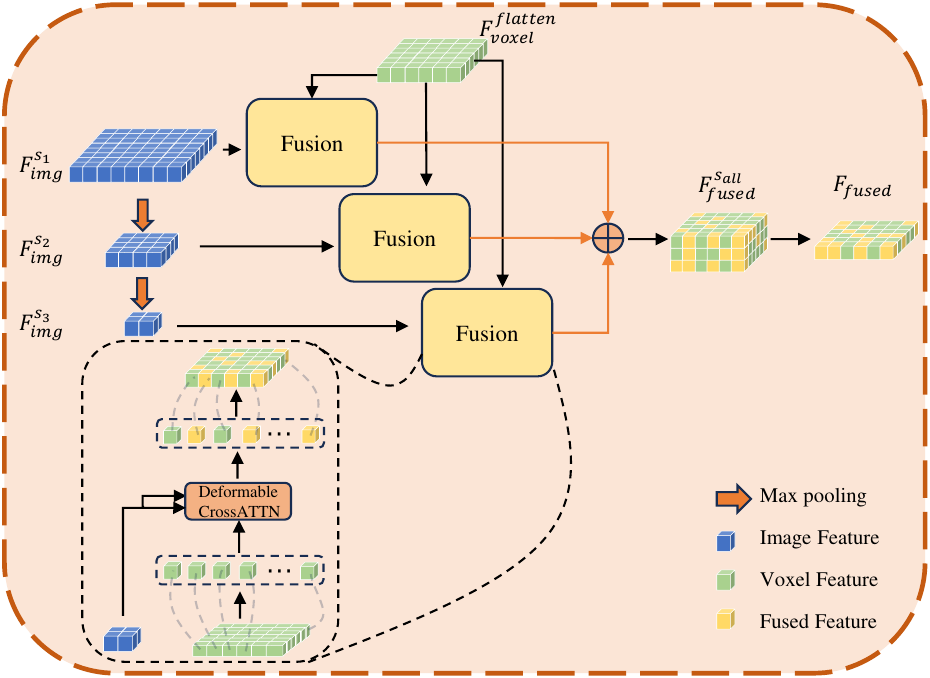}
	\caption{Structure of VWF}
	\label{fig: VWF}
\end{figure}

To address the weak perception of distant small targets by low-resolution LiDAR, we use image features to supplement the potentially missing features of distant small targets in the voxels. However, due to the sparse point clouds collected by low-resolution LiDAR, the generated voxel features may be biased. Therefore, we designed a module called the Voxel-Wise Fusion module (VWF) to supplement the features voxel by voxel. This module utilizes image features at different scales to complement voxel features, thereby achieving more accurate feature supplementation. As shown in Fig. \ref{fig: VWF}.

First, we need to obtain image features at different scales. We perform pooling operations on the original image features to obtain $F_{img}^{s_i}$,
\begin{equation}
    F_{img}^{s_i} = maxpool_i(F_{img}^{ori}),
\end{equation}
where, $F_{img}^{s_i}\in\mathbb{R}^{b\times c\times l_{i}\times w_{i}}$ represents the image features at the $i$th scale, $maxpool_i$ denotes the pooling layer used to obtain image features at the $i$th scale, and $F_{img}^{ori}\in\mathbb{R}^{b\times c\times h_c\times w_c}$ represents the original image features extracted from the raw input features.

Next, in order to process voxel features voxel by voxel, we flatten the obtained voxels,
\begin{equation}
    F_{voxel}^{flatten} = flatten(F_{voxel}^{ori}),
\end{equation}
where, $F_{voxel}^{flatten}\in\mathbb{R}^{h*l*w\times b*c}$ represents the voxel features after being flattened into a two-dimensional tensor, $F_{voxel}^{ori}\in\mathbb{R}^{h\times l\times w\times b\times c}$ represents the original voxel features. 

To effectively supplement the voxel features, we employ a mechanism similar to attention-based fusion using reference points.
This method allows each query point to only query the regions of interest, meaning attention is computed by sampling a certain number of points around the reference points. We use each voxel feature block from the flattened voxel features to query image features at different scales, obtaining updated flattened voxel features. Each voxel feature block is updated using the following formula: 
\begin{equation}
    \begin{split}
    CrossATTN(F_{voxel}^{flatten}, F_{img}^{s_i}, P) = \\ \sum\limits_{k=1}^J{{A_k^{s_i}}{W_k^{s_i}}{F_{img}^{s_i}(P+{\delta{P_k^{s_i}}})}},
    \end{split}
\end{equation}

where, $P$ represents the reference point, and $J$ denotes the number of points sampled around the reference point, ${s_i}$ represents the $i$th scale, ${A_k^{s_i}}\in\lbrack0,1\rbrack$ represents learnable attention weights, $W_{k}^{s_i}\in\mathbb{R}^{c\times c}$ represents learnable weights generated from the $F_{img}^{s_i}$, $\delta{P_k^{s_i}}$ represents the predicted offset to the reference point $P$, ${F_{img}^{s_i}(P+{\delta{P_k^{s_i}}})}$ represents the feature at location ${P+\delta{P_k^{s_i}}}$.

After fusing the image features with voxel features at all scales, we obtain the fused features for each scale, and we reshape them back to the original dimensions, obtaining $F_{fused}^{s_i}\in\mathbb{R}^{h\times l\times w\times b\times c}$. Then, we concatenate all the fused features along the channel dimension. Finally, we apply a fully connected layer to obtain the ultimate fused feature,
\begin{equation}
F_{fused}^{s_{all}} = concat(F_{fused}^{s_1}, F_{fused}^{s_2}, ..., F_{fused}^{s_i}),
\end{equation}
\begin{equation}
F_{fused} = {W_F}{F_{fused}^{s_{all}}} + {B_F},
\end{equation}
where, $F_{fused}^{s_{all}}\in\mathbb{R}^{h\times l\times w\times b\times i*c}$ represents the feature obtained by concatenating all the fused features across different scales, $F_{fused}\in\mathbb{R}^{h\times l\times w\times b\times c}$ represents the ultimate fused feature, $W_F$ and $B_F$ represent the learnable parameters in the linear layer.

\subsection{Feature Offset Correction Module}
When supplementing voxel features, image features are queried based on $J$ sampled points around the reference point. Therefore, if the LiDAR perceives an object at a certain location while the camera may fail to perceive it due to occlusion or field of view issues, the reference point will be inaccurate. Consequently, the attention scores for querying at this location become relatively lower than elsewhere. We refer to this discrepancy as feature offset.


 \begin{figure}[h]
	\centering
	\includegraphics[height=36mm]{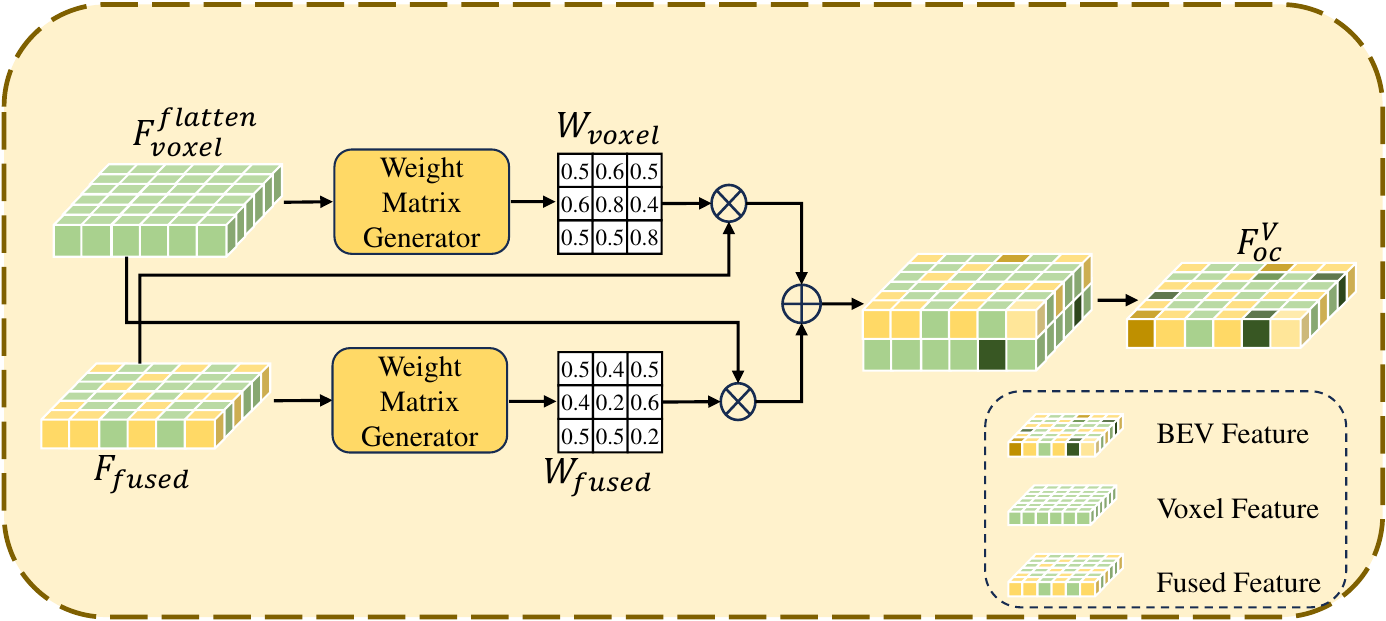}
	\caption{Structure of FOCM}
	\label{fig: focm}
\end{figure}

To address this issue, we designed the Feature Offset Correction Module (FOCM), as shown in Fig. \ref{fig: focm}. We obtain respective weight matrices, $W_{voxel}$ and $W_{fused}$, for the original voxel feature $F_{voxel}^{ori}$ and the fused voxel feature $F_{fused}$ through a weight matrix generator. The weight matrix generator consists of a fully connected layer that is used to obtain the attention scores of the feature maps. We then multiply the original voxel feature and the weight matrix for fused voxel feature, as well as the fused voxel feature and the weight matrix for the original voxel feature. This yields two feature maps with weighted scores. Subsequently, these two feature maps are concatenated and passed through a linear layer for full connection, resulting in the corrected feature $F_{oc}$. The mathematical expression is as follows:
\begin{equation}
    F_{oc}= W_{oc}(F_{voxel}W_{fused} \copyright  F_{fused}W_{voxel})+B_{oc},
\end{equation}
 where $F_{oc}$ represents the corrected feature, $W_{fused}$ denotes the feature weight matrix generated from cross-modal fusion features, $W_{voxel}$ represents the feature weight matrix generated from the original voxel features, {\copyright} denotes concatenation, and $W_{oc}$ and $B_{oc}$ represent trainable parameters in the linear layer, $W_{voxel}$ and $W_{fused}$ represent the weight matrices generated by the original voxel features and the fused voxel features, respectively.

 \subsection{Regional Feature Enhancement Algorithm}
Due to a significant reduction in the laser LiDAR's resolution at the vehicle, it results in a decrease in point cloud density, leading to the regional features becoming indistinct. Therefore, we propose the Regional Feature Enhancement Algorithm (RFEA) to address this issue. RFEA first applies Gaussian filtering to the feature map. Subsequently, we calculate the difference between each point and its surrounding points, obtaining a difference matrix with the same feature scale as the original. 
Then we assess this difference matrix and empirically set a threshold, which is determined as the optimal value through repeated experiments. When the value at a particular location exceeds this threshold, we set it to 1.
Conversely, when it is below the threshold, we set the value at that location to zero. The algorithmic process is shown in Algorithm \ref{alg:RFEA}. Afterwards, we obtain the regional difference map $M_d$$\in\left(0,1\right)$. 
\begin{algorithm}[h]
\caption{RFEA}\label{alg:RFEA}
\begin{algorithmic}[1]
\Require
feature map $F_{oc}$
\Ensure
regional difference map $M_d$ 

\State Smoothing the input feature map with Gaussian filtering.
\State Determine whether the current point $P_c$ has one or more of the neighboring points, including the top $P_t$, bottom $P_b$, left $P_l$, and right $P_r$ neighbors, and the corresponding values are $V_c, V_t,V_b, V_l, V_r$.
\State Create an empty tensor $F_{RFEA}^{empty}$ with the same size as the input feature map $F_{oc}$.
\While{not all points on $F_{oc}$ have been traversed}
\If{having neighboring points}
\State the values of the missing neighboring points are set to zero.
\State $V_d = \frac{|(V_c-V_t)|+|(V_c-V_b)|+|(V_c-V_l)|+|(V_c-V_r)|}{4}$
\ElsIf{no neighboring points}
\State $V_d = V_c$
\EndIf
\If{$V_d>V_{thres}$}
\State $V_d=1$
\ElsIf{$V_d<V_{thres}$}
\State $V_d=0$
\EndIf
\State Place $V_d$ into $F_{RFEA}^{empty}$ at the corresponding position of $P_c$.
\EndWhile
\State Obtain the regional difference map $M_d$
\end{algorithmic}
\end{algorithm}

\subsection{Collaborative Perception Module}
To address the issue of limited bandwidth while still needing to transmit large amounts of data in collaborative perception, we designed a feature selection mechanism to efficiently utilize the bandwidth.

\begin{figure}[h]
	\centering
	\includegraphics[height=35mm]{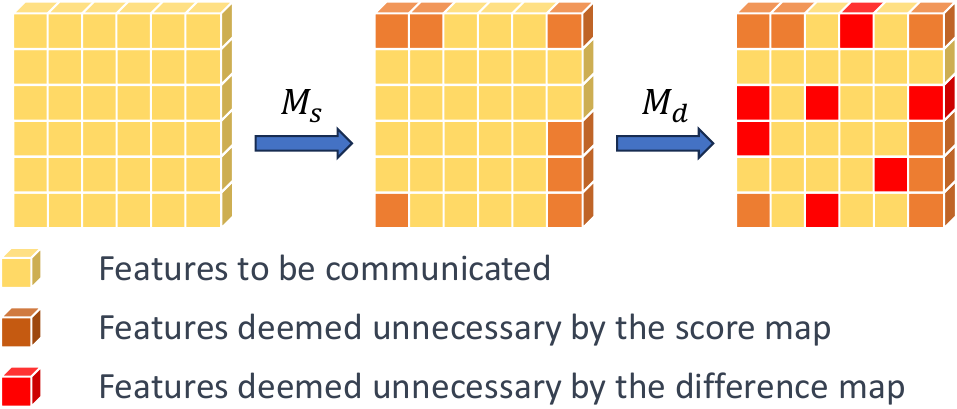}
	\caption{Regional difference map feature}
	\label{fig: difference_map_feature}
\end{figure}

\textbf{Feature selection.} First, we utilize a regional score map generator $\Phi_{gen}(\cdot)$. We input the obtained BEV feature map $F_{oc}$ into $\Phi_{gen}(\cdot)$ to generate confidence scores for each position in the feature map. Regions with confidence scores greater than a certain threshold are considered high-confidence areas, and all values in those regions are set to 1. Conversely, regions with confidence scores less than the threshold are considered low-confidence areas, and all values in those regions are set to 0.
The setting of this threshold is based on experience.
The binary map obtained as described above is referred to as the regional score map $M_s$. Then, we complement the regional difference map $M_d$ obtained from RFEA with $M_s$ to obtain the regional confidence map $M_{sd}$, which guides where the features should be transmitted, the principle of $M_{sd}$ is illustrated in Fig. \ref{fig: difference_map_feature}. 

\begin{figure}[h]
	\centering
	\includegraphics[height=48mm]{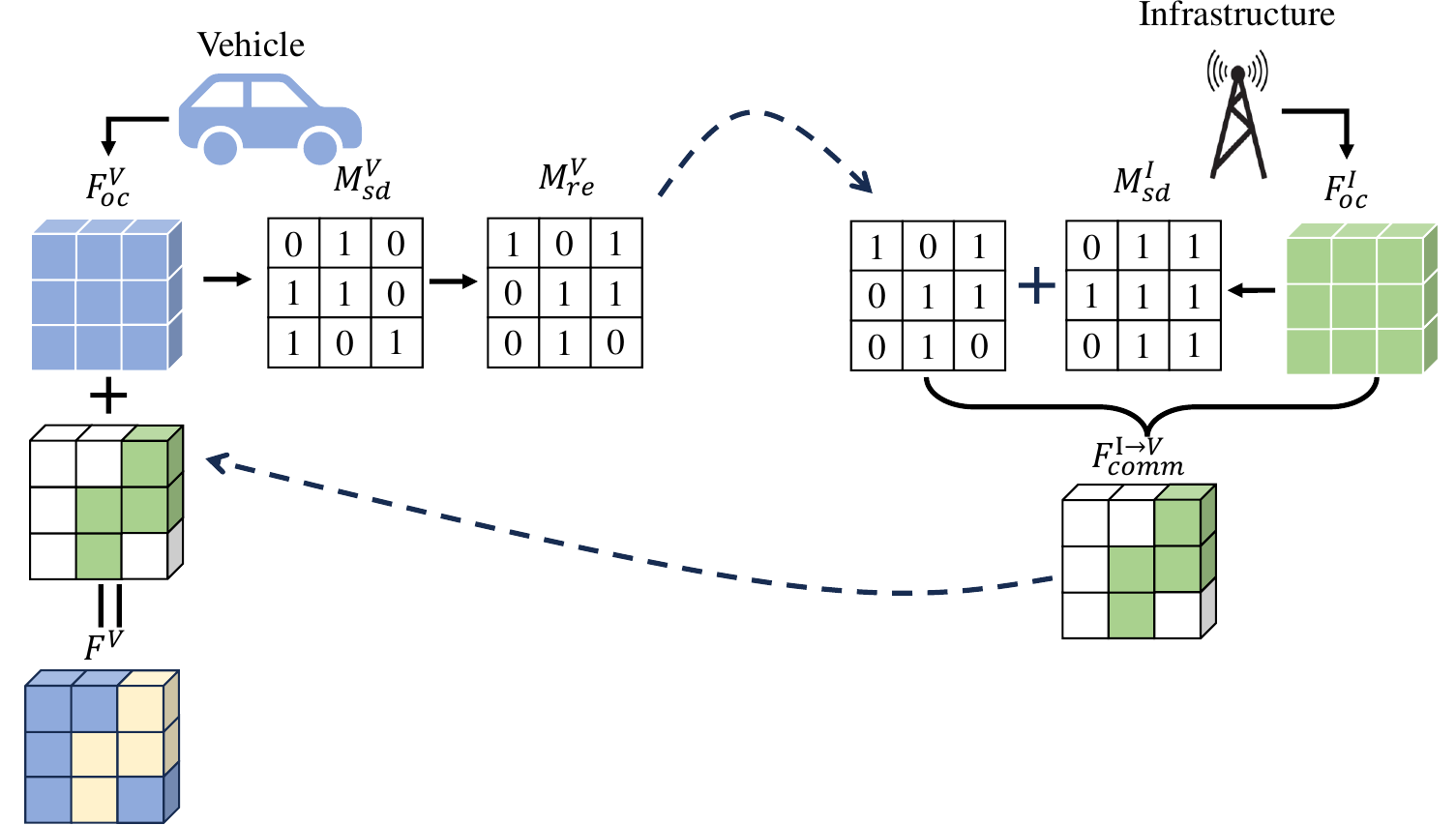}
	\caption{Diagram of cooperative perception}
	\label{fig: comm}
\end{figure}

\textbf{Communication process.} At the beginning of communication, the vehicle needs to convey its own requirements before the infrastructure can accurately assist the vehicle in perception. The overall communication process is illustrated in Fig. \ref{fig: comm}. Therefore, based on $M_{sd}$, we will utilize the following equation to obtain the request map $M_{re}$: 
\begin{equation}
    M_{re}=1-M_{sd}.
\end{equation}

During the communication process, the vehicle first sends the request map $M_{re}$ to the collaborator. The collaborator then compares the received request map $M_{re}$ with its own regional confidence map $M_{sd}$ to identify the features that can satisfy the vehicle's request for transmission:

\begin{equation}
    F_{comm}^{I \to V}=F_{oc}^I \times (M_{re}^V \odot M_{sd}^I),
\end{equation}
where $F_{comm}^{I \to V}$ represents the feature map that satisfies the perception request of the vehicle, with $I \to V$ indicating that the feature map is sent from the infrastructure to the vehicle, $F_{oc}^I$ is the unprocessed feature map at the infrastructure, $M_{re}^V$ denotes the request map of the vehicle, and $M_{sd}^I$ represents the regional confidence map at the infrastructure, $\odot$ represents performing the XNOR operation on two matrices.

\textbf{Message fusion.} After receiving $F_{comm}^{I \to V}$ from the infrastructure, it is further supplemented and fused with the local feature map $F_{oc}^V$ of the vehicle to obtain the final feature map $F^V$. The specific process is as follows:
\begin{equation}
    F^V = SA(stack(F_{comm}^{I \to V},F_{oc}^V)),
\end{equation}
where stack represents stacking the two feature maps, and $SA(\cdot)$ represents using the self-attention method to integrate the vehicle feature maps into the infrastructure's feature maps.

\textbf{Object detection.} Use the decoder to decode the features into objects,  including class and regression output.
\begin{equation}
    {\mathcal{O}^V}=\phi_{dec}(F^V).
\end{equation}
Through the aforementioned selection of transmitted features, unnecessary features for the vehicle are removed, thereby transmitting only the features of interest to the vehicle, achieving efficient utilization of communication bandwidth. 

\subsection{Loss Function}
To train the overall system, we supervise two tasks during training: the generation of the regional score map and object detection. Our regional score map generator reuses the parameters of the detection decoder. Supervised with one detection loss, the overall loss is $L=L_{det}(\mathcal{O}^V,\mathcal{O})$, where $\mathcal{O}$ is vehicle's ground-truth objects, $L_{det}$ is the detection loss \cite{DBLP}.

\section{Experiment and Analysis}
We conducted 3D object detection tasks on the DAIR-V2X dataset in the real-world scenario to validate the performance of our work. The detection results are evaluated using the Average Precision (AP) at Intersection over Union (IoU) thresholds of 0.50 and 0.70. Communication results are calculated in bytes on a logarithmic scale (base 2) for message size. To fairly compare communication results intuitively, we do not take into account any additional data/feature/model compression.
\subsection{Dataset and Experimental Setup}
DAIR-V2X \cite{9879243} is a publicly available collaborative perception dataset collected in real-world environments. The dataset consists of two main components: data captured from the perspective of intelligent vehicles, including images and point cloud data, and data captured from the perspective of infrastructure's sensors, also including images and point cloud data. The perception range covers an area of 201.6m × 80m.

The original DAIR-V2X dataset did not annotate targets outside the field of view. Therefore, we utilized the version of DAIR-V2X re-annotated by Yifan Lu and others \cite{lu2023robust}, which covers a 360-degree detection range, as our experimental dataset. We represent the field of view as a Bird's Eye View (BEV) image with dimensions of (200, 504, 64), and a precision of 0.4m per pixel for both length and width.



We compared the cooperative perception algorithms under different vehicle LiDAR precisions. Among them, V2VNet, V2X-ViT \cite{10.1007/978-3-031-19842-7_7}, and DiscoNet are unimodal algorithms with point cloud data as input. Where2comm is also a unimodal algorithm, but it supports point cloud or image data as input; here, we use point cloud data as input. BM2CP \cite{pmlr-v229-zhao23a} is a multimodal algorithm, using point cloud and image data as input for the model.


\subsection{Quantitative Evaluation}
Tables \ref{table: diffbeam} demonstrate the performance of various collaboration methods under different accuracies of vehicle's LiDAR, especially under conditions of the original resolution, half of the original resolution, and one-fourth of the original resolution. The results indicate that our method performs well under different LiDAR accuracies. Under the condition of LiDAR accuracy being the original resolution, compared with state-of-the-art methods, our performance has improved by 2.55\% in AP@0.5 and 1.27\% in AP@0.7. When the resolution is reduced by half, we achieve performance improvement by 3.18\% in AP@0.5 and 2.99\% AP@0.7. When the resolution is reduced by one-fourth, we also achieve performance improvement by 4.40\% in AP@0.5 and 3.65\% AP@0.7.

At the same time, the above data also indicates that when LiDAR accuracy decreases, other methods are significantly affected. This is shown in the figures as a sharp drop in target detection accuracy as LiDAR accuracy declines. However, our method, although affected by reduced sensor accuracy, not only maintains a lead in accuracy compared to other methods, but also does not experience a sharp decline.

We believe that when other methods reduce the accuracy of LiDAR, resulting in a sparser point cloud and insufficiently clear regional features of the target, the detection capability for distant small objects significantly decreases. In contrast, our approach addresses this issue by introducing camera assistance and enhancing the regional features of the target. This is why our method exhibits such performance.

\begin{table}[h]
\centering
	\scalebox{0.71}{
 \begin{tabular}{|c|c|c|c|c|}
    \hline
    Method & Comm & AP@0.5 & AP@0.7 \\
    \hline
    No Collaboration  & 0 & 0.5009 / 0.4892 / 0.4253 & 0.4359 / 0.4104 / 0.3997 \\  
    \hline
    V2VNet  & 24.21 & 0.5625 / 0.5511 / 0.4918 & 0.4304 / 0.4023 / 0.3877 \\ 
    \hline
    V2X-ViT  &22.62 &0.5394 / 0.5238 / 0.5008 &0.4329 / 0.4179 / 0.4084 \\
    \hline
    DiscoNet  & 22.62 & 0.5411 / 0.5299 / 0.5038 & 0.4332 / 0.4186 / 0.4037 \\ 
    \hline
    Where2comm  & 22.62 & 0.6479 / 0.6286 / 0.5981 & 0.4955 / 0.4750 / 0.4592 \\ 
    \hline
    BM2CP  &   22.62   &  \textbf{0.6762} / 0.6474 / 0.6156    &  0.5068 / 0.4901 / 0.4726      \\ 
    \hline
    LCV2I(Ours)  & 22.62 & 0.6734 / \textbf{0.6604} / \textbf{0.6421} & \textbf{0.5082} / \textbf{0.5049} / \textbf{0.4957} \\ 
    \hline
\end{tabular}
}
\caption{The performance of different methods under various vehicle's LiDAR precisions. "Comm" represents the communication bandwidth, and in AP@0.5 and AP@0.7, from left to right, the LiDAR precisions are the original resolution, half the original resolution, and a quarter of the original resolution.} 
\label{table: diffbeam}
\end{table}

\begin{figure}[h]
	\centering
	\includegraphics[height=35mm]{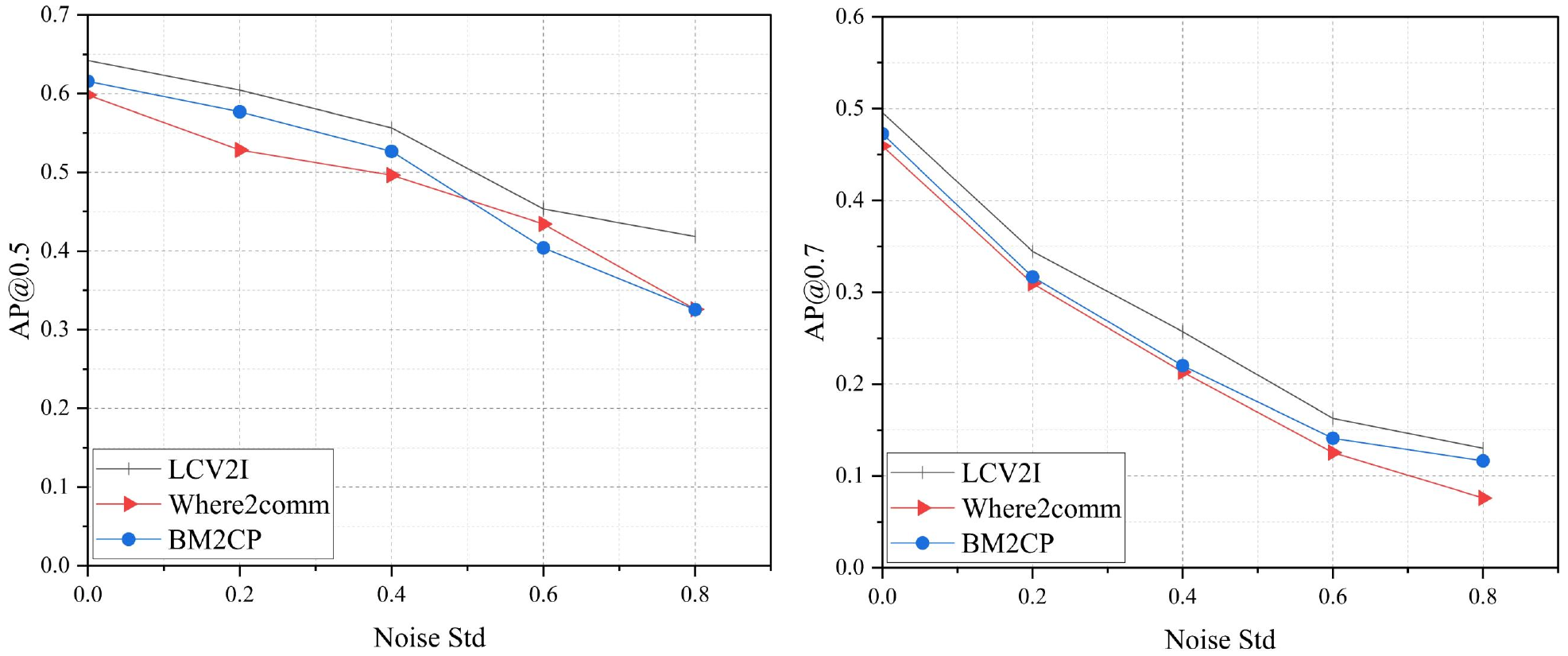}
	\caption{Robustness test with LiDAR at one-quarter of the original resolution}
	\label{fig: noise}
\end{figure}

Additionally, we introduced Gaussian noise into the network to test the robustness of LCV2I. In this experiment, we selected only BM2CP and Where2comm, whose accuracy is similar to LCV2I, for comparison. The accuracy of the other models differs significantly from LCV2I, so robustness experiments were not conducted for them. During this experiment, our LiDAR resolution is maintained at one-quarter of the original resolution, as shown in Fig. \ref{fig: noise}.

From Fig. \ref{fig: noise}, it can be seen that when Gaussian noise is introduced, the detection accuracy of all models decreases. However, when the noise standard deviation exceeds 0.4, under an IoU threshold of 0.5, the AP of LCV2I decreases more slowly compared to the other two models. Additionally, under an IoU threshold of 0.7, while the AP decrease rate of LCV2I is similar to BM2CP, it is still slower than that of Where2comm.

Therefore, compared to existing methods that rely on high-resolution LiDAR for vehicles and infrastructure, LCV2I not only reduces costs and improves detection accuracy but also offers better robustness.

\subsection{Ablation study}
As shown in Table \ref{table: ablation}, we use Where2comm as the baseline. With the original LiDAR accuracy reduced to one-fourth on the vehicle side, the baseline achieves an accuracy of 59.81\%AP@0.5 and 45.92\%AP@0.7. Introducing camera data as a supplement and using VWF to fuse data from these two sensors, we obtain accuracy gains of 2.36\%AP@0.5 and 0.18\%AP@0.7. Subsequently, by incorporating FOCM to correct the fused features, our work achieves accuracy gains of 1.10\%AP@0.5 and 2.11\%AP@0.7. Finally, building upon the aforementioned, we employ RFEA to enhance the regional features of the targets, observing optimal performance and obtaining accuracy gains of 0.94\%AP@0.5 and 1.72\%AP@0.7.

\begin{table}[h]
\centering
	\scalebox{0.8}{
 \begin{tabular}{|ccc|c|c|c|}
    \hline
    VWF  & FOCM & RFEA &LiDAR resolution & AP@0.5 & AP@0.7 \\
    \hline
     &  &  & Ori/4  &0.5981 &0.4592 \\
    \hline
    $\surd$ &  &  & Ori/4  & 0.6217 & 0.4574\\
    \hline
    $\surd$ & $\surd$ &  & Ori/4  & 0.6327 & 0.4785 \\
    \hline
    $\surd$ & $\surd$ & $\surd$ & Ori/4  & 0.6421 & 0.4957 \\
    \hline
\end{tabular}
}
\caption{Performance comparison of various approaches with the vehicle LiDAR restricted to quarter of original resolution in DAIR-V2X.}
\label{table: ablation}
\end{table}

Table \ref{table: Communication Strategies} demonstrates the impact of different communication strategies on detection performance under the same LiDAR accuracy at the vehicle end. The results indicate that when our framework's communication strategy uses only the confidence map proposed by Where2comm, the performance is 0.96\% lower compared to using both regional score map and regional difference map under the same communication bandwidth in AP@0.5. In other words, our communication strategy can achieve higher performance with the same communication bandwidth. Without using our communication strategy, achieving similar performance would require a higher communication bandwidth.

\begin{table}[h]

\centering
	\scalebox{0.8}{
 \begin{tabular}{|c|c|c|c|c|}
    \hline
    Method & LiDAR resolution & Comm & AP@0.5 & AP@0.7 \\
    \hline
    Where2comm & Ori/4 & 22.62 & 0.5981 & 0.4592 \\
    \hline
    ours' & Ori/4 & 22.62 & 0.6325 & 0.4703 \\
    \hline
    ours & Ori/4 & 22.62 & \textbf{0.6421} & \textbf{0.4957} \\
    \hline
\end{tabular}
}
\caption{Performance comparison of various approaches with the vehicle LiDAR restricted to quarter of original resolution in DAIR-V2X.}
\label{table: Communication Strategies}
\end{table}

\subsection{Qualitative evaluation}
\begin{figure}[ht]
	\centering
	\includegraphics[height=45mm]{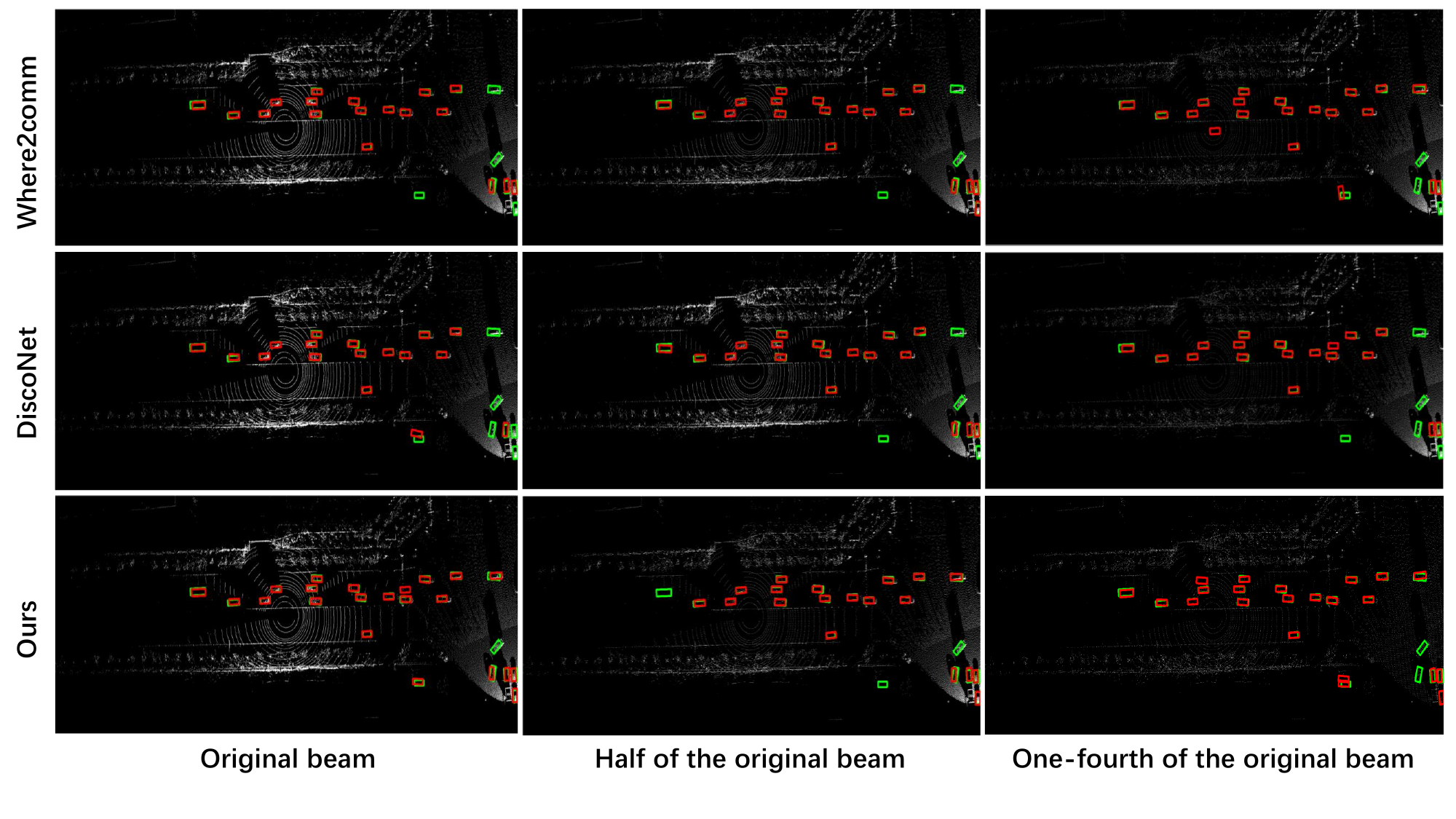} 
	\caption{Visual comparison of detection results}
	\label{fig: scene1}
\end{figure}

Finally, we conducted qualitative experiments to demonstrate the effectiveness of our work. We selected the same road section with the most complex vehicle conditions to showcase our results. We compared our algorithm with Where2Comm and DiscoNet. From Fig. \ref{fig: scene1}, it can be observed that while the other two methods also achieved good results, their detection capabilities for distant targets decreased significantly as the accuracy of the LiDAR decreased. However, our approach, which incorporates a camera sensor and utilizes region-based feature enhancement algorithms, exhibits stronger detection capabilities for distant targets even under significant reduction of the sensor's resolution. 

\begin{figure}[ht]
	\centering
	\includegraphics[height=18mm]{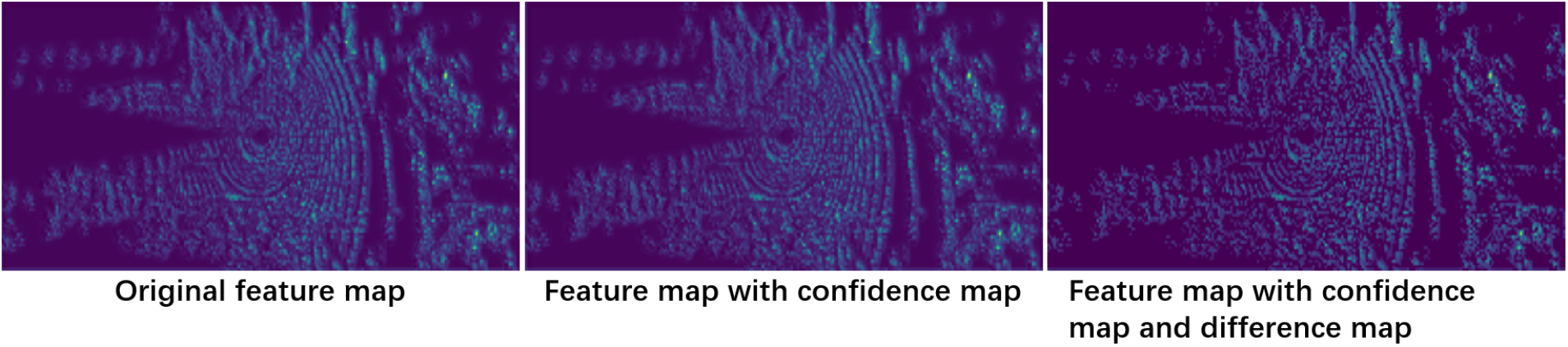} 
	\caption{Visual comparison of feature maps}
	\label{fig: feature map}
\end{figure}

Additionally, we compared the feature maps processed with the confidence map proposed by Where2comm with the feature maps processed with both regional  score map and regional difference map simultaneously. The results, as shown in the Fig. \ref{fig: feature map}, reveal that the original feature maps are relatively blurry, and the region-based features of the targets are not prominent enough. Applying the confidence map proposed by Where2Comm did not significantly improve the situation. However, when we further processed the feature maps using our REFA-generated regional difference map, it is evident that the region-based features of the targets in our method are more prominent. Moreover, irrelevant information has been removed, resulting in a relatively sparse overall feature map. This further saves transmission bandwidth.

\section{Conclusion}
In this paper, we introduce LCV2I, a framework that enhances perception by incorporating data from the vehicle's camera. Leveraging the semantic information from both the camera and LiDAR sensors under the influence of VWF, the framework enables better perception of distant small targets. To address feature misalignment caused by insufficient fusion, we propose FOCM to correct feature offset. Additionally, to tackle the issue of regional feature blurring in data collected by low-resolution LiDAR, we design RFEA. Utilizing the regional feature map obtained from RFEA, we further optimize the communication strategy. Our framework, using low-resolution sensors, achieves high detection performance, significantly reducing costs. Experimental results on the DAIR-V2X dataset demonstrate that our proposed LCV2I achieves higher detection accuracy and lower deployment costs compared to existing algorithms. 


\bibliography{aaai25}

\end{document}